\documentclass[final]{cvpr}

\usepackage{graphicx}
\usepackage{amsmath}
\usepackage{amssymb}
\usepackage{booktabs}
\usepackage{bm}
\usepackage{multirow}
\usepackage[export]{adjustbox}
\usepackage{times}
\usepackage{epsfig}

\usepackage{comment}
\usepackage{color}
\usepackage{xspace}

\usepackage[pagebackref,breaklinks,colorlinks]{hyperref}

\def\cvprPaperID{9296} 

\def\confYear{2022}

\begin{document}

\title{im2nerf: Image to Neural Radiance Field in the Wild}
\author{
  \hspace{-1.3cm}
  \begin{tabular}[t]{c}
    Lu Mi$^{1,2,\dagger}$, Abhijit Kundu$^1$, David Ross$^1$, Frank Dellaert$^{1,4}$, Noah Snavely$^{1,3}$,  Alireza Fathi$^1$\\
    $^1$Google Research $^2$MIT, $^3$Cornell Tech, Cornell University $^4$Georgia Tech\\
\end{tabular}
}

\maketitle

\pagestyle{empty}
\thispagestyle{empty}

\let\thefootnote\relax\footnotetext{\leftline{$\dagger$Work done during internship at Google.}}
\let\thefootnote\relax\footnotetext{\leftline{Corresponding to \texttt{lumi@mit.edu}}}

\begin{abstract}
We propose im2nerf, a learning framework that predicts a continuous neural object representation given a single input image in the wild, supervised by only segmentation output from off-the-shelf recognition methods. The standard approach to constructing neural radiance fields \cite{Mildenhall20eccv_nerf} takes advantage of multi-view consistency and requires many calibrated views of a scene, a requirement that cannot be satisfied when learning on large-scale image data in the wild. We take a step towards addressing this shortcoming by introducing a model that encodes the input image into a disentangled object representation that contains a code for object shape, a code for object appearance, and an estimated camera pose from which the object image is captured. Our model conditions a NeRF on the predicted object representation and uses volume rendering to generate images from novel views. We train the model end-to-end on a large collection of input images. As the model is only provided with single-view images, the problem is highly under-constrained. Therefore, in addition to using a reconstruction loss on the synthesized input view, we use an auxiliary adversarial loss on the novel rendered views. Furthermore, we leverage object symmetry and cycle camera pose consistency. We conduct extensive quantitative and qualitative experiments on the ShapeNet dataset \cite{choy20163d} as well as qualitative experiments on Open Images dataset \cite{kuznetsova2020open}. We show that in all cases, im2nerf achieves the state-of-the-art performance for novel view synthesis from a single-view unposed image in the wild.

\end{abstract}

\section{Introduction}
\begin{figure*}[t]
\centering
\includegraphics[width=1.0\linewidth]{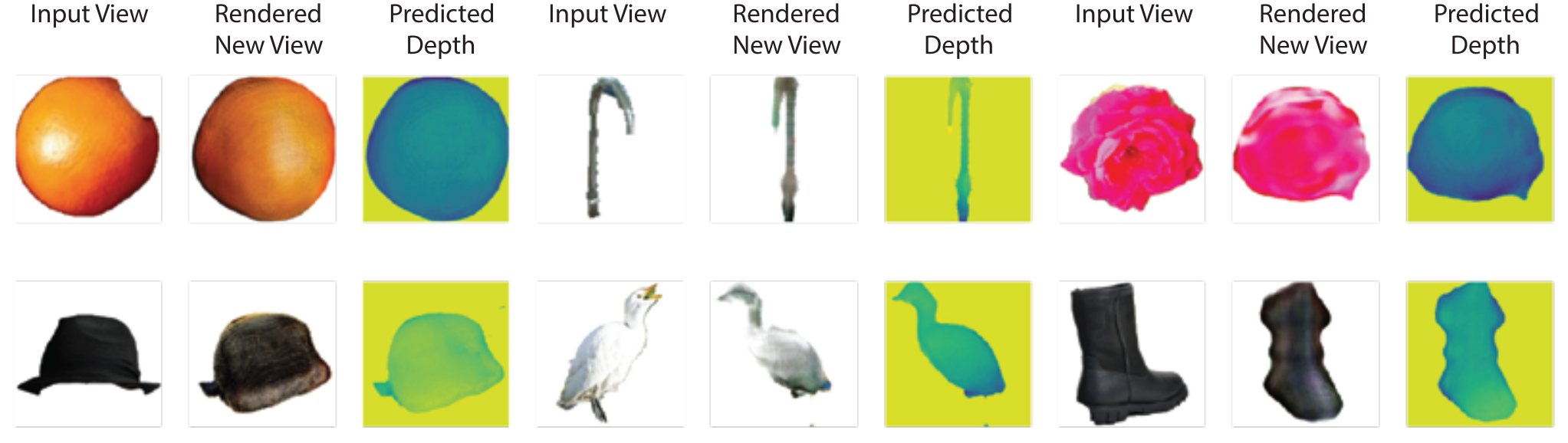}
	 \caption{We propose im2nerf, a learning framework that predicts a continuous neural object representation given a single-view unposed input image in the wild.}
\label{fig:teaser}
\end{figure*}

In recent years, the computer vision community has made great strides towards learning accurate representation of the 3D shape and appearance of objects and scenes~\cite{Mildenhall20eccv_nerf,Jang21iccv_CodeNeRF,Niemeyer21cvpr_GIRAFFE,Chan21cvpr_piGAN,Lin21iccv_BARF}. Learning such representations can result in more realistic view synthesis and image generation~\cite{Wang21cvpr_IBRNet,reizenstein2021common}, as well as richer 3D scene understanding~\cite{Zhi21iccv_SemanticNeRF,Niemeyer21cvpr_GIRAFFE}. To learn 3D representations, prior work often takes advantage of multi-view consistency, given multiple images of the scene or object from different views with known camera poses. However, the assumption that the model can have access to multiple posed views of scenes either during training or testing does not hold for large-scale image data in the wild.

A few recent approaches aim at predicting 3D shape and appearance of objects from unposed single-view images \cite{ye2021shelf,Goel20eccv_wo_keypoint}. However, 
the 3D object representations used (e.g., meshes or voxels combined with implicit representations) result in reconstructions that lack detail. Furthermore, the problem is often constrained by encouraging object instances to look similar to an average category template which results in a low level of deformation in the final predictions.

In this work, we propose \textbf{im2nerf}, a learning framework that maps a given single input image to a predicted continuous neural object representation that through volume rendering synthesizes novel views of the object. Our approach is based on an auto-encoder model. The encoder decomposes each input image in terms of object geometry, object appearance and camera pose. The decoder is a NeRF-like~\cite{Mildenhall20eccv_nerf,Jang21iccv_CodeNeRF} model that can render the input image or new views of the object. It generates rays corresponding to pixels in the image based on a given camera pose, and uses a multi-layer perceptron (MLP) conditioned on the predicted shape and appearance representations to predict the volume density and view-dependent radiance for each 3D point.

As we lack 3D supervision or multiple views of the in-the-wild images, inspired by some recent approaches \cite{ye2021shelf,Goel20eccv_wo_keypoint,Wu20cvpr_symmetry,Jain21iccv_DietNeRF}, we leverage other sources of meta-supervision present in our knowledge of the 3D world. Our encoder-decoder approach encourages the object geometry and appearance to be consistent with the input image through differentiable rendering. Furthermore, we use an adversarial loss to encourage the rendering of novel views of the object to look realistic. As the novel views are rendered conditioned on the same representation as the input view, multi-view consistency is forced across all the rendered images. We furthermore leverage the fact that many object categories have in principle a symmetric 3D shape. Finally, we propose a cycle consistency loss that indirectly constrains the estimated camera poses to be geometrically meaningful. We feed the rendered novel view back to the encoder to re-estimate the camera pose.

To summarize, our contributions are as follows:
\begin{itemize}
    \item We propose an unsupervised learning framework that predicts a continuous 3D neural representation given a single input image in the wild.
    \item We approach this highly under-constrained problem setting by leveraging auxiliary sources of meta-supervision including an adversarial loss on rendered novel views, using scene box, encouraging object symmetry and using cycle camera pose consistency.
    \item Our results achieve the state-of-the-art performance for novel view synthesis from a single-view unposed image in the wild.
    \item We further apply our method to a weakly supervised setting where we are provided with pose supervision only for a very small subset of the data. We show that our method is able to leverage the unlabeled data to significantly improve the model to achieve a comparable performance to a model trained with full supervision of pose annotations.
\end{itemize}

\section{Related Work}

\noindent \textbf{Neural Representations.} Using implicit functions to represent the 3D geometry and appearance of objects and scenes has gained popularity in recent years~\cite{Park19cvpr_deepsdf,Mescheder19cvpr_occupancy_net,Najibi20cvpr_dops,Mildenhall20eccv_nerf,Jang21iccv_CodeNeRF,nguyen2019hologan,Chan21cvpr_piGAN,Yu21cvpr_pixelNeRF,Wang21cvpr_IBRNet,Niemeyer21cvpr_GIRAFFE,Lin21iccv_BARF,Jain21iccv_DietNeRF,Boss21iccv_NeRD,MartinBrualla21cvpr_nerfw}. These methods use an MLP as a function that maps the continuous spatial coordinates and view angles into signed distance, occupancy, color or density values. The advances in implicit representations together with the recent progress in differentiable rendering~\cite{kato18cvpr_mesh_renderer,li18tog_differentiable_monte_carlo,Jiang20cvpr_sdfdiff,Liu19iccv_soft_rasterizer,Niemeyer20cvpr_DVR}, have led to the emergence of NeRF~\cite{Mildenhall20eccv_nerf}, which captures viewpoint-varying image properties by storing volume density and view-dependent radiance fields from multi-view observations. 

NeRF makes several assumptions: (a) the camera poses of input images and their intrinsics are known, (b) the model is optimized on a single static scene, and (c) many input images from multiple views are required to achieve the best results. Various extensions to NeRF have made progress on relaxing some of these assumptions. PixelNeRF~\cite{Yu21cvpr_pixelNeRF}, GRF~\cite{Trevithick21iccv_GRF} and IBRNet~\cite{Wang21cvpr_IBRNet} can generalize to more than one scene by using an image encoder to condition the neural radiance field on image features. These methods still require multiple posed views of the scene during the training stage but can use a small number of views during the inference stage. iNeRF~\cite{YenChen20iros_iNeRF}, NeRF{-}{-}~\cite{Wang21arxiv_NeRFminusminus}, GNeRF~\cite{meng2021gnerf} and BARF~\cite{Lin21iccv_BARF} aim at relaxing the known camera pose condition. However, these methods work best when the initial camera pose is close to the correct one. Henzler \etal~\cite{Henzler21cvpr_object_categories} and Li \etal~\cite{Li21arxiv_n3dvs} learn a 3D object or scene representation from video. NeRF-VAE~\cite{Kosiorek21icml_NeRF_VAE}, GIRAFFE~\cite{Niemeyer21cvpr_GIRAFFE}, Stelzner \etal~\cite{Stelzner21arxiv_ObSuRF} and DeVries \etal~\cite{DeVries21iccv_GSN} learn to decompose the scene into objects or/and generate scenes with conditional radiance fields. CodeNeRF~\cite{Jang21iccv_CodeNeRF} uses an auto-decoder approach (similar to DeepSDF~\cite{Park19cvpr_deepsdf}) that jointly learns a code representation per object and the weights of the conditional NeRF multi-layer perceptron. All of these methods assume they are provided with multiple posed views of the objects or the scenes during the training stage, and do not show results on large-scale images in the wild.

\medskip
\noindent \textbf{Supervised Single-view 3D Reconstruction.} 
Various data collection efforts have played a key role in early progress in 3D reconstruction from 2D images: synthetic datasets like ShapeNet~\cite{Wu15cvpr_3d_shapenet} and SunCG~\cite{Song17cvpr_suncg}, generation of synthetic data from real-world images~\cite{Chen163dv,Zuffi19iccv_safari,Varol17cvpr_synthetic_humans}, as well as other work that annotate real world images by fitting 3D CAD models to object silhouettes like Pascal3D~\cite{Xiang14wacv_pascal3d} and Pix3D~\cite{Sun18cvpr_pix3d}.
Some of the earlier methods leverage such data to formulate the problem assuming full 3D supervision~\cite{Wu15cvpr_3d_shapenet,choy20163d,Wu16nips_3d_gan,Rematas20cvpr_Neural_Voxel_Renderer,Gkioxari19iccv_mesh_rcnn,Girdhar16eccv,Tulsiani18cvpr_factoring_shape,Groueix18cvpr_AtlasNet,Mescheder19cvpr_occupancy_net}. However, requiring ground-truth 3D supervision restricts these methods to synthetic or small datasets, since ground-truth 3D shapes are costly or impossible to acquire at large scale.

\medskip
\noindent \textbf{Unsupervised Single-view 3D Reconstruction.}
Progress in differentiable renderers~\cite{kato18cvpr_mesh_renderer,li18tog_differentiable_monte_carlo,Jiang20cvpr_sdfdiff,Liu19iccv_soft_rasterizer,Niemeyer20cvpr_DVR} has allowed building end-to-end methods that lift an image into a 3D representation and render the 3D representation back to a novel view. DeepSDF~\cite{Park19cvpr_deepsdf} optimizes an implicit signed distance function that is used for representing the object in 3D by rendering it back to multiple images, requiring only 2D supervision. Goel \etal~\cite{Goel20eccv_wo_keypoint} predict a mesh-based 3D representations of objects from 2D images. However, the mesh-based representation often lacks the expressiveness needed to render high fidelity images.  Wu \etal~\cite{Wu20cvpr_symmetry} explore using symmetry as a constraint to encourage more accurate 3D representations.
PrGAN~\cite{gadelha20173d}, HoloGAN~\cite{nguyen2019hologan}, GRAF~\cite{Schwarz20neurips_graf}, pi-GAN~\cite{Chan21cvpr_piGAN} and GIRAFFE~\cite{Niemeyer21cvpr_GIRAFFE} use a generative model that is trained with an adversarial loss for learning 3D representation of objects from unposed images. These models are unconditional, meaning that they are able to generate an object or a scene given a random noise vector. However, they wouldn't necessarily be able to generate such representation conditioned on an input image.

The closest approach to ours is the work by Ye \etal~\cite{ye2021shelf}, which proposes a method for learning 3D object representations in the wild. They use a combination of voxel and implicit representations to predict the 3D shape and appearance of the objects. However, the reconstructions usually lack details as a result of choices made in the 3D representation. 

\section{Method}

\begin{figure}[t]
\centering
\includegraphics[width=1.0\linewidth]{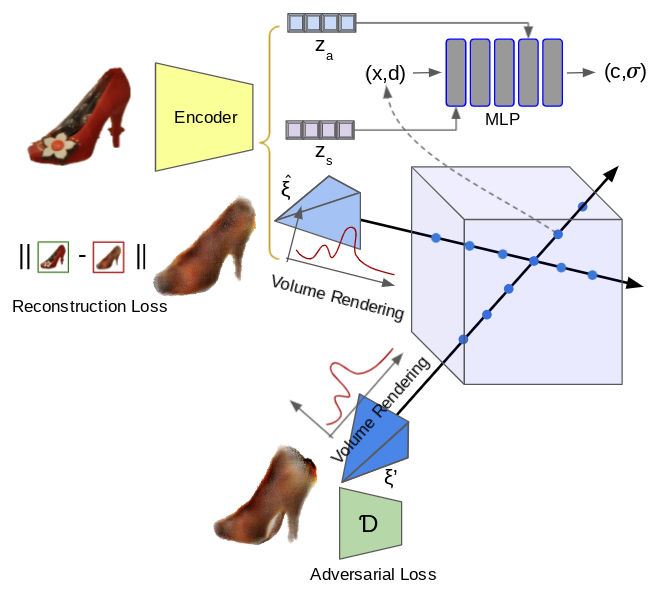}
	 \caption{\textbf{Overview of our method.} Given an input image, the encoder predicts a shape  $z_s$ and an appearance code $z_a$ and estimates the pose of the camera $\hat{\xi}$ that captures the input image. The decoder conditions a NeRF on the predicted shape and appearance representations and uses volume rendering to generate images from novel views. In addition to using a photometric reconstruction loss for input view, we apply an adversarial loss on rendered images from novel views. In addition, we further constrain the problem by using a scene box, cycle camera pose consistency and object symmetry (for symmetric object categories).}
\label{fig:model}
\end{figure}

An overview of our method is shown in Figure \ref{fig:model}. Our goal is to learn a model that maps an input image segmented with an object mask to a 3D representation that can be rendered from novel views. Since the model should generalize to images in the wild, we do not assume any other sources of 2D or 3D supervision during training or testing. Our model consists of an encoder-decoder architecture. The encoder takes in the masked image of an object and predicts a latent shape and a latent appearance code, as well as the pose of the camera that captured the image (assuming the object is at the origin in its canonical frame and the camera intrinsics are known). The decoder takes in the latent codes and a camera pose (with known intrinsics) and renders a novel view image. 

The key challenge in our problem setting is that we must optimize the parameters of the encoder and the conditional NeRF MLP in the decoder from a collection of unposed single-view images. Inspired by earlier work~\cite{ye2021shelf,Goel20eccv_wo_keypoint,Chan21cvpr_piGAN,Niemeyer21cvpr_GIRAFFE} in this space, we propose to use meta-supervision that is present in our knowledge of the 3D world. We combine a differentiable rendering loss on the input view with an adversarial loss on the novel views, together with the geometric consistency that is inherent in the conditional NeRF to train our model. Furthermore, we leverage other real-world knowledge such as scene box, object symmetry and cycle camera pose consistency loss to optimize our model.

\subsection{Model}

Our model follows a typical encoder-decoder structure, and the details of model components are introduced as follows,

\textbf{Encoder.} Our encoder consists of two feature extractor towers followed by prediction heads. The first tower predicts the shape and appearance latent codes for the object. The MLP in the decoder is conditioned on these codes to predict the color and density of the points in 3D from different view angles. The second tower predicts the pose of the camera that captures the input image. We assume the camera intrinsics are known and the world coordinate system is centred on the 3D object. The camera pose is described by a rotation and a translation matrix. To predict the rotation matrix, our network predicts 4 parameters: $\cos{\theta}$, $\sin{\theta}$, $\cos{\phi}$, $\sin{\phi}$ where $\theta$ is {\it azimuth} and $\phi$ is {\it elevation}. We then
compute the rotation matrix as 
$\bm{R} = \bm{R_\theta} \bm{R_\phi}$. 
Our camera pose tower predicts the translation of the object in the camera frame by predicting $t_x$, $t_y$ and $t_z$. Since the camera is always looking at the object, $t_z$ is always positive.

\textbf{Decoder.} The decoder receives three inputs---a camera pose, a shape code and an appearance code---and outputs a rendered image. 
It first generates a set of pixel rays based on the input camera pose. Points are sampled along the rays, and for each point, color and density values are estimated conditioned on the shape and appearance codes.

{\it Conditioning on Shape and Appearance Codes.} Inspired by the  recent progress on generative NeRF models \cite{Chan21cvpr_piGAN,Jang21iccv_CodeNeRF,Kosiorek21icml_NeRF_VAE}, we use a multi-layer perceptron that maps scene coordinates $\bm{x}$ and view directions $\bm{d}$ to their corresponding density $\bm{\hat{\sigma}}$ and RGB color values $\bm{\hat{c}}$ conditioned on the shape and appearance codes
\begin{equation}
    (\bm{\hat{c}}, \bm{\hat{\sigma}}) = F(\gamma(\bm{x}), \gamma(\bm{d}), \bm{z_a}, \bm{z_s}),
\end{equation}
where $\bm{z_s}$ is the shape code, $\bm{z_a}$ is the appearance code, and $\gamma$ is the positional encoding function to capture high frequency position and view direction details. In our architecture design, the point density $\bm{\hat{\sigma}}$ is predicted from the positional encoding $\gamma(\bm{x})$ conditioned on the shape code $\bm{z_s}$, while the RGB color $\bm{\hat{c}}$ additionally depends on the view direction encoding $\gamma(\bm{d})$ and the appearance code $\bm{z_a}$.

{\it Volumetric Rendering.} Given camera extrinsics $\bm{\xi}$, we sample pixel rays $\bm{r}(t) = \bm{o} + t\bm{d}$ given their origin $\bm{o}$ and direction $\bm{d}$. We then use the volume rendering operator to calculate the expected color $\bm{c}(\bm{r})$ of the ray $\bm{r}$ from the predicted densities and colors $(\sigma_i, c_i)$ of the sampled points along the ray~\cite{Mildenhall20eccv_nerf}
\begin{equation}
\begin{aligned}
\bm{\hat{c}}(\bm{r}) = \sum_{i=1}^NT_i(1-\text{exp}(-\sigma_i\delta_i))c_i, \\ \text{where}~T_i=\text{exp}\Big(-\sum_{j=1}^{i-1}\sigma_j\delta_j\Big)
\end{aligned}
\end{equation}
where $t_n$ and $t_f$ represent the near and far bounds, $N$ is the number of evenly-spaced bins between the bounds and $\delta_i = t_{i+1}-t_i$ is the distance of adjacent samples. We predict a soft object foreground mask per ray as $\alpha = 1- \text{exp}(-\sigma_i\delta_i)$. We follow NeRF~\cite{Mildenhall20eccv_nerf} to use hierarchical sampling, and calculate the final rendered colors $\bm{\hat{c}}(\bm{r})$ using coarse and fine points sampled from rays.

\subsection{Training}

\label{sec:training}

As we lack 3D supervision or multiple views of the in-the-wild images, inspired by some recent approaches~\cite{ye2021shelf,Chan21cvpr_piGAN,Goel20eccv_wo_keypoint}, we leverage other sources of meta-supervision present in our knowledge of the 3D world.

\textbf{Input view reconstruction loss.} Given the input image, the encoder predicts the 3D object shape and appearance representation, and estimates the pose of the camera that captures the input image. The decoder conditions a NeRF MLP on the shape and appearance code and uses volume rendering to render an image given a novel view camera pose. We use the decoder to render back the input view by providing it with the predicted camera pose and object representation. In addition to rendering the color image, we render soft object segmentation masks represented by the alpha value $\bm{\hat{\alpha}}$ in volume rendering. Therefore, for both image reconstruction loss $\mathcal{L}_\textrm{recon}^c$ and mask (alpha) reconstruction loss $\mathcal{L}_\textrm{recon}^{\alpha}$, we calculate the pixel-wise $L_2$ loss 
\begin{equation}
\mathcal{L}_\textrm{recon}^c = \|\bm{c} - \bm{\hat{c}}\|_2
\end{equation}
\begin{equation}
\mathcal{L}_\textrm{recon}^{\alpha} = \|\bm{\alpha} - \bm{\hat{\alpha}}\|_2
\end{equation}

\textbf{View synthesis adversarial loss.} Our method is not provided with novel view images of the objects. To avoid predicting a degenarate 3D representation that only looks like an object from the input view, we utilize adversarial training with a discriminator to force realistic novel view synthesis. During training, a novel view pose $\bm{\xi'}$ is randomly sampled from a pre-defined prior distribution $p(\bm{\xi})$. Then the new view image and alpha mask are rendered and fed together with random input images to a discriminator that ideally shouldn't be able to tell which image is real and which one is fake. 

One can think of our model as a generator $G^c$ that receives an image $\bm{c}$, input object mask $\bm{\alpha}$ and a novel view camera pose ${\bm{\xi'}}$ as inputs and generates an image $\bm{\hat{c}'}$ as output (internally, there is an encoder inside this generator that encodes the input image into shape and appearance codes, and estimates its camera pose, and a decoder that renders the novel view image conditioned on the predicted codes). We direct the generated images to a traditional convolutional discriminator $D^c$ for adversarial training. We use the following adversarial loss:

\begin{equation}
\begin{aligned}
\mathcal{L}_\textrm{adv}^c &= \text{log}D^c(\bm{c}) + \text{log}(1-D^c(G^c(\bm{c},\bm{\alpha},\bm{\xi'})))
\end{aligned}
\end{equation}

Similarly, we use a generator $G^\alpha$ that receives the same inputs but outputs the soft foreground object mask, for which we use a discriminator $D^\alpha$ that tries to distinguish predicted mask from the input image masks. We use the following adversarial loss for the predicted $\alpha$ foreground mask:

\begin{equation}
\begin{aligned}
\mathcal{L}_\textrm{adv}^\alpha &= \text{log}D^\alpha(\bm{\alpha}) + \text{log}(1-D^\alpha(G^\alpha(\bm{c},\bm{\alpha},\bm{\xi'})))
\end{aligned}
\end{equation}

\textbf{Pose consistency loss.} Our model consists of an encoder and a decoder. The encoder takes in the input image and predicts shape and appearance codes and estimates the pose of the input view camera. Conditioned on the object shape and appearance codes and a randomly sampled camera pose $\bm{\xi'}$, the decoder renders a novel view $\bm{\hat{c'}}$. We propose a cycle consistency loss that indirectly constrains the estimated camera poses to be geometrically meaningful. We feed the rendered novel view $\bm{\hat{c'}}$ back to the encoder to re-estimate the camera pose $\bm{\hat{\xi'}}$. We formulate pose consistency loss as 

\begin{equation}
\mathcal{L}_\textrm{poc} = \|\bm{\xi'}-\bm{\hat{\xi'}}\|_2
\end{equation}

\textbf{Encouraging symmetry.} 
For the categories with symmetric objects, we further force the object shape to be symmetric across the $y$ axis for the objects that have this property. We do this by feeding $\bm{x}=(x_1, |x_2|, x_3)$ as the input to the neural radiance field MLP. That this only forces symmetry on the object shape. Object appearance additionally takes in the ray direction which lets it to be more flexible.

\textbf{Pose supervised loss.} In our ablation studies, we also study the effect of weakly supervision and full supervision from given pose ground truth, using an additional camera pose estimation loss. The camera pose estimation loss consists of two parts: a mean squared error loss for translation and a mean squared error loss that is applied to the predicted angles in the rotation matrix. Note that such loss is not applied in the unsupervised setting.

\section{Experiments}

\begin{table}[t]
\centering
\setlength{\tabcolsep}{2.0pt}
\small
\begin{tabular}{cc|cc|cc|cc}
\toprule
\multicolumn{2}{c|}{Supervision} & \multicolumn{2}{c|}{chair} & \multicolumn{2}{c|}{car} & \multicolumn{2}{c}{airplane} \\ \hline
\multicolumn{1}{c|}{Labeled} & \multicolumn{1}{c|}{Unlabeled} & \multicolumn{1}{c}{PSNR} & \multicolumn{1}{c|}{SSIM} & \multicolumn{1}{c}{PSNR} & \multicolumn{1}{c|}{SSIM} & \multicolumn{1}{c}{PSNR} & \multicolumn{1}{c}{SSIM} \\ \hline
\multicolumn{1}{c|}{100\%} & {0\%}  & 18.9 & 0.77 & 20.1 & 0.83 & 25.2 & 0.93 \\ 
\multicolumn{1}{c|}{1\%} & 99\% & 18.6 & 0.77 & 19.5 & 0.82 & 24.8 & 0.92 \\ 
\multicolumn{1}{c|}{1\%} & 0\% & 17.1 & 0.72 & 19.5 & 0.81 & 23.7 & 0.90 \\
\multicolumn{1}{c|}{0\%} & 100\% & 16.4 & 0.72 &18.6  & 0.79 & 22.2 & 0.87 \\ 

\bottomrule
\end{tabular}
\caption{We perform an ablation study by training our model with different levels of supervision on chair, car, airplane in ShapeNet 3D-R2N2 \cite{choy20163d}. This study demonstrates that using only $1\%$ of pose ground truth under the weakly supervised case, and utilizing unlabeled images, leads to performance comparable to fully supervision ($100\%$) of pose.}
\label{table:shapenet ablation weak supervise}
\end{table}

\begin{figure}[t]
\centering
\includegraphics[width=1.0\linewidth]{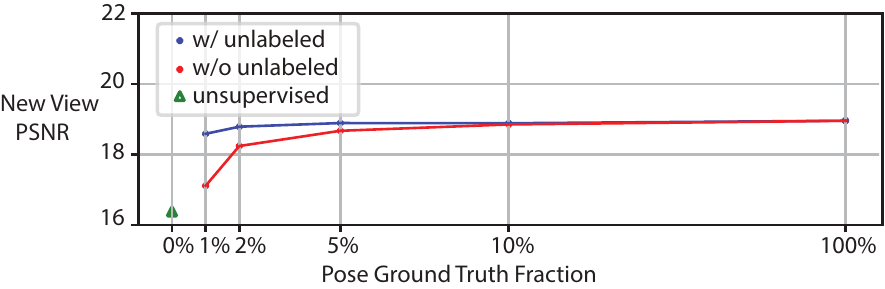}\
	 \caption{We perform an ablation study by training our model with different levels of pose supervision on ShapeNet 3D-R2N2 \cite{choy20163d} chair. Surprisingly, we find using only $1\%$ of pose ground truth under the weakly supervised case, and utilizing unlabeled images, leads to the performance comparable to full supervision ($100\%$).}
\label{fig:shapenet weak supervised chair}
\end{figure}

We show quantitative and qualitative results on the ShapeNet 3D-R2N2 \cite{choy20163d} dataset. ShapeNet is a synthetic dataset that lets us validate the effectiveness of our method quantitatively and compare it to previous baselines. We present ablation studies that demonstrate the challenges in learning 3D object representations given only single-view input images with unknown camera poses. Furthermore, in this ablation study, we show the key role that the auxiliary losses and design choices play in helping our model achieve competitive results to a fully-supervised pose setting. Finally, we show qualitative results on the Open Images dataset ~\cite{kuznetsova2020open} that demonstrate our method's capability to learn the 3D presentation of objects and synthesize them from novel views in a large-scale realistic setup in the wild.

\subsection{Datasets}
We describe the two public benchmark datasets that the models are evaluated on in the following,

\begin{figure*}[t]
\centering

\includegraphics[width=0.95\linewidth]{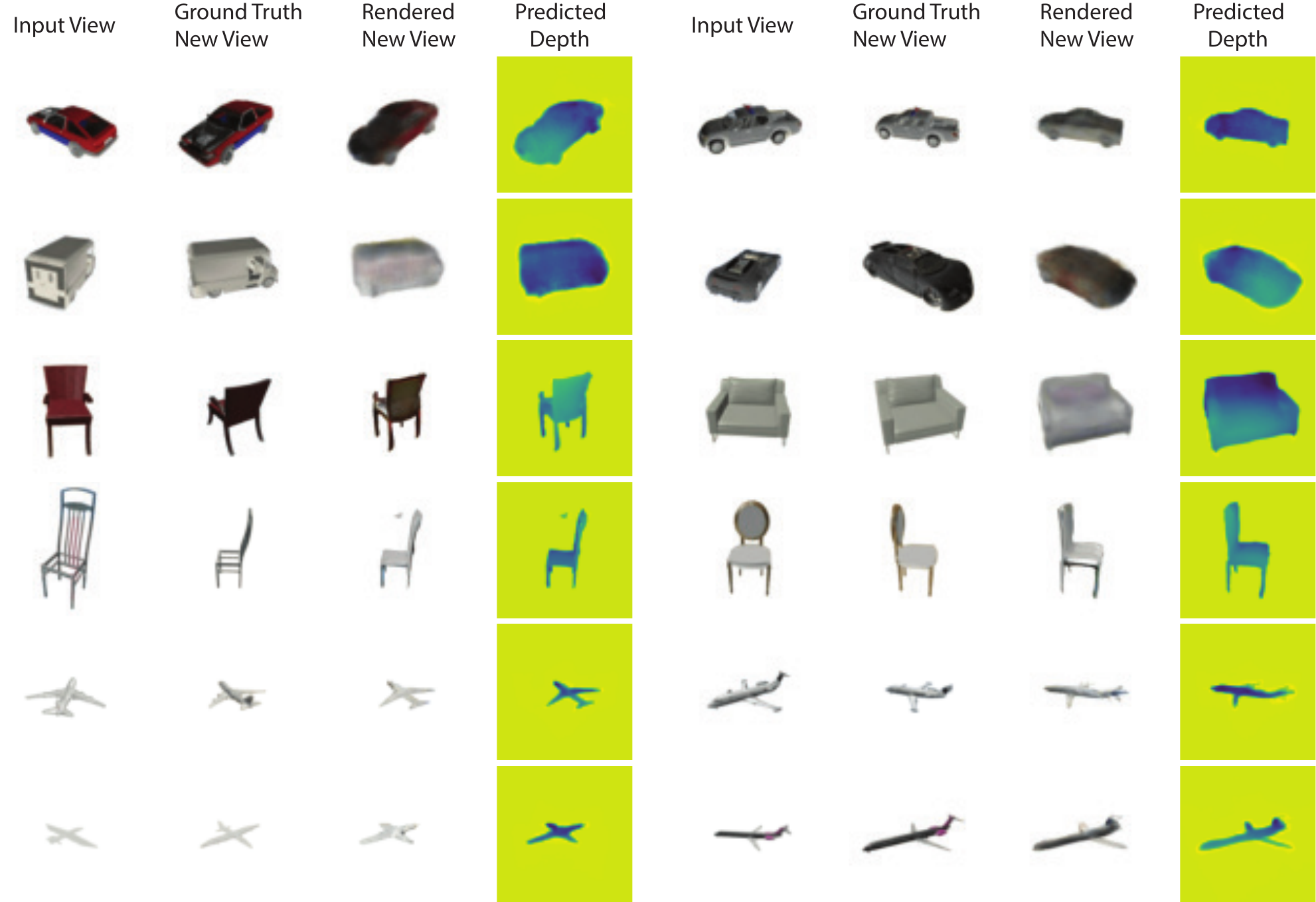}
	 \caption{We show our method's results under the unsupervised setting for chairs, cars and airplanes of the ShapeNet 3D-R2N2 dataset~\cite{choy20163d}.}
\label{fig:shapenet}
\end{figure*}

\textbf{ShapeNet 3D-R2N2}~\cite{choy20163d}. We evaluate our method on three categories including car, chair and airplane. ShapeNet contains 24 views of each object. Since the dataset is synthetic, we could design different sets of ablation studies with providing our model with different levels of supervision, or using the ground truth views to measure the effectiveness of different design choices and losses.

\textbf{Open Images}~\cite{kuznetsova2020open}. Following Ye \etal~\cite{ye2021shelf}, we qualitatively evaluate our method on Open Images dataset.
The number of the images in each object category range from 500 to 20k. As presented in \cite{Benenson19cvpr}, the foreground masks are from annotation and are filtered by a fine-tuned occlusion classifier. In our experiments, we crop a tight bounding box around the masked object and use that as the input to our model. In Figure~\ref{fig:teaser} and Figure~\ref{fig:open_images_all}, we show qualitative results on Open Images. We evaluate our model on a wide range of object categories and show that it can learn to represent different types of objects including thin (water tap, guitar, saxophone), flat (hat), round (orange, balloon), deformable (birds, animals), and objects with complicated textures (Christmas trees).

\subsection{Experiment Setup}
Here we describe the model architecture, training setup and evaluation setup.

\textbf{Model Architecture.} We use an input imagesize of $112 \times  112$. The latent code dimension for both the appearance code $\bm{z_a}$ and shape code $\bm{z_s}$ is 64. For both ShapeNet and Open Images datasets, we use pre-defined camera intrinsics (the one used in ShapeNet dataset). For the conditional MLP, we use a depth of 6 with a width of 128. For each ray we use 64 coarse samples and 128 fine samples. We use a near depth of 0.1 and a far depth of 4.0 for the rays. Meanwhile, for ShapeNet dataset, we use a 3D bounding box with the size of $(-0.4, 0.4)$ in $x, y, z$ dimension. The backbone for the encoder is ResNet50 model pre-trained on ImageNet~\cite{krizhevsky2012imagenet}. The backbone for the discriminator consists of 3 fully-convolutional layers with leakyReLU activations and a dropout layer before the last fully connected layer. Since rendering $112\times 112$ rays exceeds the maximum memory capacity of a GPU, we use a patch discriminator. we stride the image to  sample patches of rays during rendering. We use render a sampled novel view patches of size $80 \times 80$ for each input image.

\textbf{Training Setup.} During training, the model takes in a single-view masked image as input, and renders back patches for the estimated input view and a randomly sampled novel view. The model is trained with various losses as described in Section~\ref{sec:training}. We use weight $1.0$ for input view reconstruction loss $\mathcal{L}_\textrm{recon}^c$ and $\mathcal{L}_\textrm{recon}^{\alpha}$. We also use weight $1.0$ for novel view synthesis adversarial loss $\mathcal{L}_\textrm{adv}^c$ and $\mathcal{L}_\textrm{adv}^{\alpha}$. We use weight $50.0$ for pose consistency loss $\mathcal{L}_\textrm{poc}$. 

We use $10$ frequencies for positional encoding $\gamma$ to encourage higher details in rendered images. We find that optimizing pose estimation with positional encoding is difficult. Therefore, similar to BARF~\cite{Lin21iccv_BARF}, we gradually increase the weight of the positional encoding from $0$ to $1$. Our model is trained for 200 epochs with the Adam optimizer with an initial learning rate of $1^{-3}$ and a exponential learning rate decay with a rate of 0.96. We train our model on 8 Tesla Nvidia V100 GPUs with a batch size of $8$. 

\textbf{Evaluation Setup.} Since our model is not provided with ground-truth canonical pose of the objects, the predicted camera poses could be rotated with respect to the ground-truth. This is still a valid prediction since the object representation is also rotated with respect to the ground-truth canonical pose. To evaluate the accuracy of our rendered novel view images quantitatively, we use a universal offset matrix during evaluation that maps the predicted poses to the ground-truth ones.

\subsection{Ablation Study}

\begin{table}[t]
\centering
\begin{tabular}{ccccc}
\toprule
constraint & \multicolumn{1}{c}{recon.} & \multicolumn{1}{c}{+adv.} & \begin{tabular}[c]{@{}l@{}}+scene\\ box\end{tabular} & \multicolumn{1}{c}{\begin{tabular}[c]{@{}c@{}}+pose.\\ consistency\end{tabular}}\\ \hline
\multicolumn{1}{c|}{PSNR} & 12.7 & 12.8 &16.3 & \textbf{16.4} \\
\multicolumn{1}{c|}{SSIM} & 0.59 & 0.64 &0.71  & \textbf{0.72} \\ \bottomrule
\end{tabular}
\caption{We perform an ablation study of losses and design choices such as adversarial loss, scene bounding box, cycle camera pose consistency loss for the unsupervised setting on ShapeNet 3D-R2N2 chair \cite{choy20163d}. One additional loss to the objective function or constraint is added at a time (from left to right).}
\label{tab: ablation shapenet 3 categories}
\end{table}

\begin{figure}[t]
\centering
\includegraphics[width=1.0\linewidth]{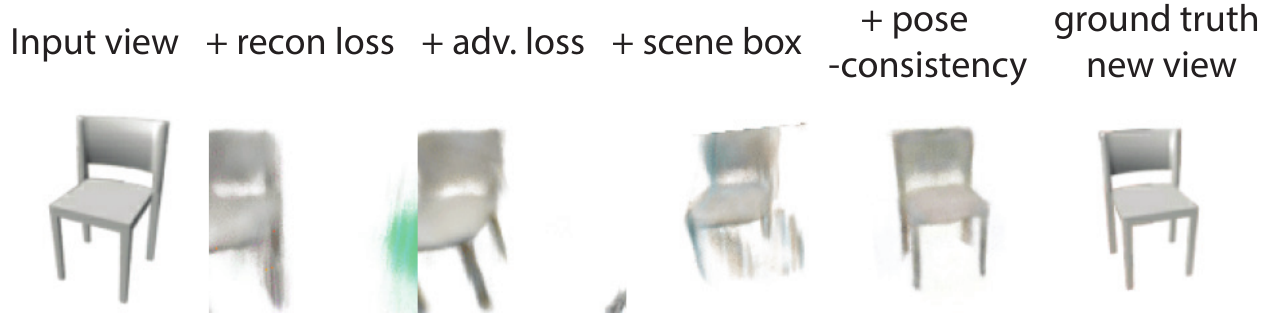}\
	 \caption{We perform an ablation study of losses and design choices such as adversarial loss, scene bounding box, cycle camera pose consistency loss for the unsupervised setting on ShapeNet 3D-R2N2 chair \cite{choy20163d}. One additional loss to the objective function or constraint is added at a time (from left to right).}
\label{fig:shapenet ablation chair}
\end{figure}

As shown in Figure~\ref{fig:shapenet weak supervised chair} and Table~\ref{table:shapenet ablation weak supervise}, we inspect our model under different levels of pose ground truth supervision (with single-view input). We show that our model by using only $1\%$ of posed data and leveraging the other $99\%$ of unlabeled data, could achieve results comparable to a fully supervised setting, and outperform training without unlabeled data.

Meanwhile, we also perform studies with various combinations of losses and design choices for the unsupervised condition: adversarial loss $\mathcal{L}_{\text{adv}}$, scene box, and pose consistency loss $\mathcal{L}_\textrm{poc}$. We find that using photometric reconstruction loss ($\mathcal{L}_\textrm{recon}^c$  only does not converge in our problem setup where we only have access to single-view images with unknown camera poses. We demonstrate the key role each of the following losses or design choices make in enabling our method achieve competitive results: (1) adversarial loss for images $\mathcal{L}_\textrm{adv}^c$ and masks $\mathcal{L}_\textrm{adv}^{\alpha}$, (2) scene box, and (3) cycle camera pose consistency loss $\mathcal{L}_\textrm{poc}$. Results are shown in Figure~\ref{fig:shapenet ablation chair} and Table~\ref{tab: ablation shapenet 3 categories}. Furthermore, we show qualitative results in other categories including car and airplane from ShapeNet dataset in Figure~\ref{fig:shapenet}.

\subsection{Evaluation on Benchmark Datasets}

\textbf{Results on ShapeNet 3D-R2N2.} Meanwhile, we compare our model im2nerf with other baselines in the unsupervised settings using single-view and unknown camera pose, our proposed model im2nerf achieves significantly better results in comparison to the baselines on the ShapeNet 3D-R2N2 dataset \cite{choy20163d}, as shown in Table~\ref{tab: shapenet quantitative}.

\begin{itemize}
    \item {\it Shelf-Supervised}. We compare our method to that of Ye \etal~\cite{ye2021shelf}. They use a combination of voxel and implicit representations to predict the 3D shape and appearance of the objects. We refer to this model as \textit{shelf-supervised} in the our results. In order to make the comparison, We use their publicly released source code to perform the evaluations on both ShapeNet 3D-R2N2 and Open Images datasets. For Open Images dataset, we use their released pre-trained model for each category to render the images.
    \item {\it PrGAN}. We compare our method to that of Gadelha \etal~\cite{gadelha20173d} which we call PrGAN throughout the paper. We use the open source implementation by Ye \etal~\cite{ye2021shelf} to evaluate PrGAN on our datasets.
    \item {\it HoloGAN}. We compare our method to that of Nguyen \etal~\cite{nguyen2019hologan}. HoloGAN is added with a function that maps the volumetric feature to occupancy using reprojection consistency with foreground masks. We use the open source implementation by Ye \etal~\cite{ye2021shelf} to evaluate HoloGAN on our datasets.

\end{itemize}

\begin{figure}[t]
\centering
\includegraphics[width=0.75\linewidth]{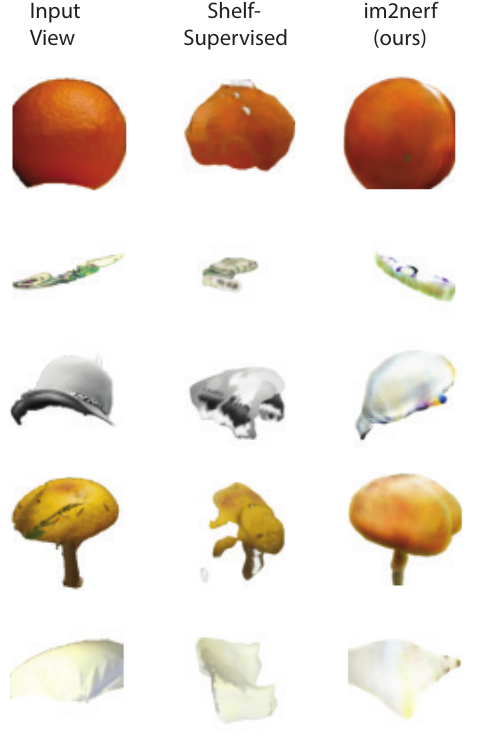}
	 \caption{Our method im2nerf's qualitative results on Open Images dataset \cite{kuznetsova2020open}. The synthesized novel views by our algorithm look more realistic than those of the Ye \etal~\cite{ye2021shelf}.}
\label{fig:open_baselines}
\end{figure}

\begin{figure*}[t]
\centering
\includegraphics[width=1.0\linewidth]{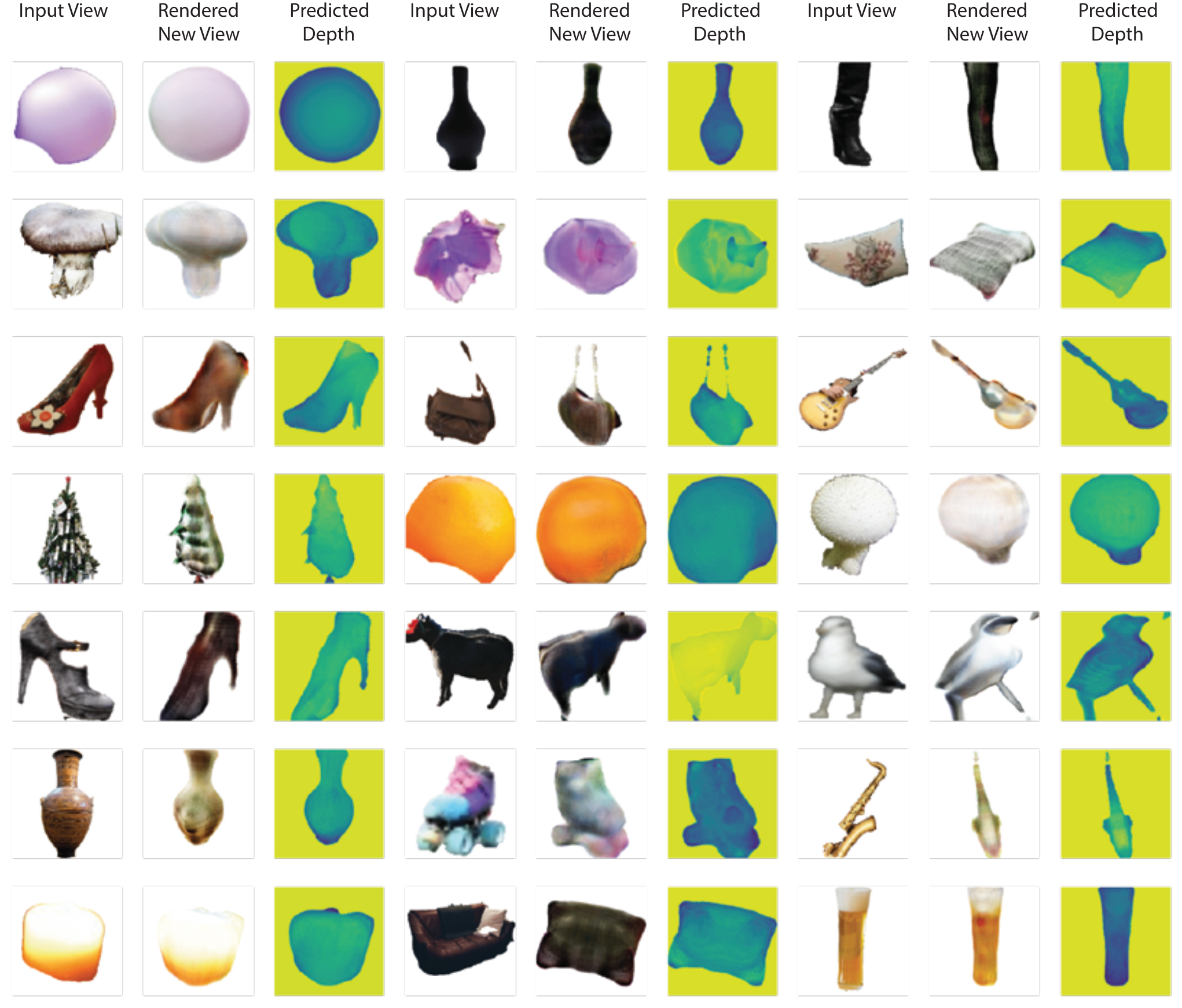}
	 \caption{Results of im2nerf model evaluated on Open Images dataset~\cite{kuznetsova2020open,Benenson19cvpr}. We show three columns of results, where in each column, from left to right we show the input image, synthesized novel view, and predicted depth of the synthesized novel view.}
\label{fig:open_images_all}
\end{figure*}

\textbf{Results on Open Images dataset.} We show qualitative results on Open Images dataset. In Figure~\ref{fig:open_baselines}, we compare our results qualitatively with that of Ye \etal~\cite{ye2021shelf}. In Figure~\ref{fig:teaser} and Figure \ref{fig:open_images_all}, we show qualitative results of our proposed method im2nerf on Open Images dataset.

\begin{table}[t]
\centering
\setlength{\tabcolsep}{2.9pt}
\begin{tabular}{c|cccc}
\toprule
\multirow{2}{*}{model} & \multirow{2}{*}{HoloGAN} & \multirow{2}{*}{PrGAN} & \multirow{2}{*}{Shelf-Supervised} & \multirow{2}{*}{\begin{tabular}[c]{@{}c@{}}im2nerf \\ (ours)\end{tabular}} \\

                       &                        &                          &                                   &                                 \\
                       \hline
chair                  & 13.2                  & 13.8                    & 13.9                             & \textbf{16.4}                           \\
car                    & 16.2                  & 16.7                    & 16.9                             & \textbf{18.6}                           \\
airplane               & 16.0                  & 16.3                    & 16.4                             & \textbf{22.2} \\ 
\bottomrule
\end{tabular}
\caption{We compare our method quantitatively with baselines including Shelf-Supervised~\cite{ye2021shelf}, HoloGAN~\cite{nguyen2019hologan}, PrGAN~\cite{gadelha20173d} evaluated on the unsupervised condition for ShapeNet~\cite{choy20163d} chair, car, and airplane. We use the average PSNR of the synthesized novel view images as the comparison metric.}
\label{tab: shapenet quantitative}
\end{table}

\section{Conclusion and Limitations}

In this paper, we present im2nerf, a learning framework that can decompose an image of an object into a 3D shape code, an appearance code, and a camera viewpoint. Our model conditions a NeRF on the 3D shape and appearance codes and uses volume rendering to synthesize novel views of the object. We exploit the meta-supervision that is present in our knowledge of the 3D world to to solve this highly under-constrained problem. In particular, in addition to using a reconstruction loss on the synthesized input view, we use an auxiliary adversarial loss on rendered novel view, leverage the fact that most objects are symmetric, and apply a scene box and cycle pose consistency loss. Interestingly, we find using only $1\%$ of pose ground truth under the weakly supervised case, and utilizing unlabeled images, leads to performance comparable to full supervision. We will release code and checkpoints upon acceptance.

While we achieve encouraging results across both synthetic and large-scale real-world datasets, our approach has several limitations. Conditional NeRF consumes a significant amount of memory and computation, which limits our ability to render high-resolution novel view images during training. Therefore, we apply the adversarial loss on coarse patches rendered from the novel view. We believe this causes our model to learn a less detailed 3D representation. Additionally, since we are using single-view images, the model may have trouble learning concave structures. To solve this issue, we believe either the model should be provided with a pseudo-depth supervision or alternatively another auxiliary self-supervised loss that would encourage the model to learn shape from shading.

{\small
\bibliographystyle{ieee_fullname}
\bibliography{main}

\begin{thebibliography}{10}\itemsep=-1pt

\bibitem{Benenson19cvpr}
Rodrigo Benenson, Stefan Popov, and Vittorio Ferrari.
\newblock Large-scale interactive object segmentation with human annotators.
\newblock In {\em CVPR}, 2019.

\bibitem{Boss21iccv_NeRD}
Mark Boss, Raphael Braun, Varun Jampani, Jonathan Barron, Ce Liu, and Hendrik
  Lensch.
\newblock {NeRD: Neural Reflectance Decomposition from Image Collections}.
\newblock In {\em ICCV}, October 2021.

\bibitem{Chan21cvpr_piGAN}
Eric Chan, Marco Monteiro, Peter Kellnhofer, Jiajun Wu, and Gordon Wetzstein.
\newblock {pi-GAN: Periodic Implicit Generative Adversarial Networks for
  3D-Aware Image Synthesis}.
\newblock In {\em CVPR}, 2021.

\bibitem{Chen163dv}
Wenzheng Chen, Huan Wang, Yangyan Li, Hao Su, Zhenhua Wang, Changhe Tu, Dani
  Lischinski, Daniel Cohen-Or, and Baoquan Chen.
\newblock Synthesizing training images for boosting human 3d pose estimation.
\newblock In {\em International Conference on 3D Vision (3DV)}, 2016.

\bibitem{choy20163d}
Christopher~B Choy, Danfei Xu, JunYoung Gwak, Kevin Chen, and Silvio Savarese.
\newblock 3d-r2n2: A unified approach for single and multi-view 3d object
  reconstruction.
\newblock In {\em Proceedings of the European Conference on Computer Vision
  ({ECCV})}, 2016.

\bibitem{DeVries21iccv_GSN}
Terrance DeVries.
\newblock {Unconstrained Scene Generation With Locally Conditioned Radiance
  Fields}.
\newblock In {\em ICCV}, October 2021.

\bibitem{gadelha20173d}
Matheus Gadelha, Subhransu Maji, and Rui Wang.
\newblock 3d shape induction from 2d views of multiple objects.
\newblock In {\em 2017 International Conference on 3D Vision (3DV)}, pages
  402--411. IEEE, 2017.

\bibitem{Girdhar16eccv}
Rohit Girdhar, David~F. Fouhey, Mikel Rodriguez, and Abhinav Gupta.
\newblock Learning a predictable and generative vector representation for
  objects.
\newblock In {\em ECCV}, 2016.

\bibitem{Gkioxari19iccv_mesh_rcnn}
Georgia Gkioxari, Jitendra Malik, and Justin Johnson.
\newblock Mesh r-cnn.
\newblock In {\em ICCV}, 2019.

\bibitem{Goel20eccv_wo_keypoint}
Shubham Goel, Angjoo Kanazawa, and Jitendra Malik.
\newblock Shape and viewpoint without keypoints.
\newblock In {\em ECCV}, 2020.

\bibitem{Groueix18cvpr_AtlasNet}
Thibault Groueix, Matthew Fisher, Vladimir~G. Kim, Bryan~C. Russell, and
  Mathieu Aubry.
\newblock {AtlasNet: A Papier-Mâché Approach to Learning 3D Surface
  Generation}.
\newblock In {\em CVPR}, 2018.

\bibitem{Henzler21cvpr_object_categories}
Philipp Henzler.
\newblock {Unsupervised Learning of 3D Object Categories from Videos in the
  Wild}.
\newblock In {\em CVPR}, 2021.

\bibitem{Jain21iccv_DietNeRF}
Ajay Jain, Matthew Tancik, and Pieter Abbeel.
\newblock {Putting NeRF on a Diet: Semantically Consistent Few-Shot View
  Synthesis}.
\newblock In {\em ICCV}, October 2021.

\bibitem{Jang21iccv_CodeNeRF}
Wongbong Jang and Lourdes Agapito.
\newblock {CodeNeRF: Disentangled Neural Radiance Fields for Object
  Categories}.
\newblock In {\em ICCV}, October 2021.

\bibitem{Jiang20cvpr_sdfdiff}
Yue Jiang, Dantong Ji, Zhizhong Han, and Matthias Zwicker.
\newblock Sdfdiff: Differentiable rendering of signed distance fields for 3d
  shape optimization.
\newblock In {\em CVPR}, 2020.

\bibitem{kato18cvpr_mesh_renderer}
Hiroharu Kato, Yoshitaka Ushiku, and Tatsuya Harada.
\newblock Neural 3d mesh renderer.
\newblock In {\em CVPR}, 2018.

\bibitem{Kosiorek21icml_NeRF_VAE}
Adam Kosiorek, Heiko Strathmann, Daniel Zoran, Pol Moreno, Rosalia Schneider,
  Sona Mokrá, and Danilo Rezende.
\newblock {NeRF-VAE: A Geometry Aware 3D Scene Generative Model}.
\newblock In {\em ICML}, 2021.

\bibitem{krizhevsky2012imagenet}
Alex Krizhevsky, Ilya Sutskever, and Geoffrey~E Hinton.
\newblock Imagenet classification with deep convolutional neural networks.
\newblock {\em Advances in neural information processing systems},
  25:1097--1105, 2012.

\bibitem{kuznetsova2020open}
Alina Kuznetsova, Hassan Rom, Neil Alldrin, Jasper Uijlings, Ivan Krasin, Jordi
  Pont-Tuset, Shahab Kamali, Stefan Popov, Matteo Malloci, Alexander
  Kolesnikov, et~al.
\newblock The open images dataset v4.
\newblock {\em International Journal of Computer Vision}, 128(7):1956--1981,
  2020.

\bibitem{Li21arxiv_n3dvs}
Tianye Li, Mira Slavcheva, Michael Zollhoefer, Simon Green, Christoph Lassner,
  Changil Kim, Tanner Schmidt, Steven Lovegrove, Michael Goesele, and Zhaoyang
  Lv.
\newblock Neural 3d video synthesis.
\newblock In {\em arXiv:2103.02597}, 2021.

\bibitem{li18tog_differentiable_monte_carlo}
Tzu-Mao Li, Miika Aittala, Fredo Durand, and Jaakko Lehtinen.
\newblock Differentiable monte carlo ray tracing through edge sampling.
\newblock In {\em ACM TOG}, 2018.

\bibitem{Lin21iccv_BARF}
Chen-Hsuan Lin, Wei-Chiu Ma, Antonio Torralba, and Simon Lucey.
\newblock {BARF: Bundle-Adjusting Neural Radiance Fields}.
\newblock In {\em ICCV}, October 2021.

\bibitem{Liu19iccv_soft_rasterizer}
Shichen Liu, Tianye Li, Weikai Chen, , and Hao Li.
\newblock Soft rasterizer: A differentiable renderer for image-based 3d
  reasoning.
\newblock In {\em ICCV}, 2019.

\bibitem{MartinBrualla21cvpr_nerfw}
Ricardo Martin-Brualla, Noha Radwan, Mehdi Sajjadi, Jonathan Barron, Alexey
  Dosovitskiy, and Daniel Duckworth.
\newblock {NeRF in the Wild: Neural Radiance Fields for Unconstrained Photo
  Collections}.
\newblock In {\em CVPR}, 2021.

\bibitem{meng2021gnerf}
Quan Meng, Anpei Chen, Haimin Luo, Minye Wu, Hao Su, Lan Xu, Xuming He, and
  Jingyi Yu.
\newblock Gnerf: Gan-based neural radiance field without posed camera.
\newblock {\em arXiv preprint arXiv:2103.15606}, 2021.

\bibitem{Mescheder19cvpr_occupancy_net}
Lars Mescheder, Michael Oechsle, Michael Niemeyer, Sebastian Nowozin, and
  Andreas Geiger.
\newblock {Occupancy Networks: Learning 3D Reconstruction in Function Space}.
\newblock In {\em CVPR}, 2019.

\bibitem{Mildenhall20eccv_nerf}
Ben Mildenhall, Pratul Srinivasan, Matthew Tancik, Jonathan Barron, Ravi
  Ramamoorthi, and Ren Ng.
\newblock {NeRF: Representing Scenes as Neural Radiance Fields for View
  Synthesis}.
\newblock In {\em ECCV}, pages 405--421. Springer, 2020.

\bibitem{Najibi20cvpr_dops}
Mahyar Najibi, Guangda Lai, Abhijit Kundu, Zhichao Lu, Vivek Rathod, Thomas
  Funkhouser, Caroline Pantofaru, David Ross, Larry~S. Davis, and Alireza
  Fathi.
\newblock Dops: Learning to detect 3d objects and predict their 3d shapes.
\newblock In {\em CVPR}, 2020.

\bibitem{nguyen2019hologan}
Thu Nguyen-Phuoc, Chuan Li, Lucas Theis, Christian Richardt, and Yong-Liang
  Yang.
\newblock Hologan: Unsupervised learning of 3d representations from natural
  images.
\newblock In {\em Proceedings of the IEEE/CVF International Conference on
  Computer Vision}, pages 7588--7597, 2019.

\bibitem{Niemeyer21cvpr_GIRAFFE}
Michael Niemeyer and Andreas Geiger.
\newblock {GIRAFFE: Representing Scenes as Compositional Generative Neural
  Feature Fields}.
\newblock In {\em CVPR}, 2021.

\bibitem{Niemeyer20cvpr_DVR}
Michael Niemeyer, Lars Mescheder, Michael Oechsle, and Andreas Geiger.
\newblock {Differentiable Volumetric Rendering: Learning Implicit 3D
  Representations without 3D Supervision}.
\newblock In {\em CVPR}, 2020.

\bibitem{Park19cvpr_deepsdf}
Jeong~Joon Park, Pete Florence, Julian Straub, Richard Newcombe, and Steven
  Lovegrove.
\newblock {DeepSDF: Learning Continuous Signed Distance Functions for Shape
  Representation}.
\newblock In {\em CVPR}, 2019.

\bibitem{reizenstein2021common}
Jeremy Reizenstein, Roman Shapovalov, Philipp Henzler, Luca Sbordone, Patrick
  Labatut, and David Novotny.
\newblock Common objects in 3d: Large-scale learning and evaluation of
  real-life 3d category reconstruction.
\newblock In {\em Proceedings of the IEEE/CVF International Conference on
  Computer Vision}, pages 10901--10911, 2021.

\bibitem{Rematas20cvpr_Neural_Voxel_Renderer}
Konstantinos Rematas and Vittorio Ferrari.
\newblock {Neural Voxel Renderer: Learning an Accurate and Controllable
  Rendering Tool}.
\newblock In {\em CVPR}, 2020.

\bibitem{Schwarz20neurips_graf}
Katja Schwarz, Yiyi Liao, Michael Niemeyer, and Andreas Geiger.
\newblock {GRAF: Generative Radiance Fields for 3D-Aware Image Synthesis}.
\newblock In {\em Adv. Neural Inform. Process. Syst.}, 2020.

\bibitem{Song17cvpr_suncg}
Shuran Song, Fisher Yu, Andy Zeng, Angel~X. Chang, Manolis Savva, and Thomas
  Funkhouser.
\newblock Semantic scene completion from a single depth image.
\newblock In {\em CVPR}, 2017.

\bibitem{Stelzner21arxiv_ObSuRF}
Karl Stelzner, Kristian Kersting, and Adam Kosiorek.
\newblock {Decomposing 3D Scenes into Objects via Unsupervised Volume
  Segmentation}.
\newblock {\em https://arxiv.org/abs/}, 2021.

\bibitem{Sun18cvpr_pix3d}
Xingyuan Sun, Jiajun Wu, Xiuming Zhang, Zhoutong Zhang, Chengkai Zhang, Tianfan
  Xue, Joshua~B. Tenenbaum, and William~T. Freeman.
\newblock Pix3d: Dataset and methods for single-image 3d shape modeling.
\newblock In {\em CVPR}, 2018.

\bibitem{Trevithick21iccv_GRF}
Alex Trevithick and Bo Yang.
\newblock {GRF: Learning a General Radiance Field for 3D Scene Representation
  and Rendering}.
\newblock In {\em ICCV}, October 2021.

\bibitem{Tulsiani18cvpr_factoring_shape}
Shubham Tulsiani, Saurabh Gupta, David Fouhey, Alexei~A. Efros, and Jitendra
  Malik.
\newblock Factoring shape, pose, and layout from the 2d image of a 3d scene.
\newblock In {\em CVPR}, 2018.

\bibitem{Varol17cvpr_synthetic_humans}
Gül Varol, Javier Romero, Xavier Martin, Naureen Mahmood, Michael~J. Black,
  Ivan Laptev, and Cordelia Schmid.
\newblock Learning from synthetic humans.
\newblock In {\em CVPR}, 2017.

\bibitem{Wang21cvpr_IBRNet}
Qianqian Wang, Zhicheng Wang, Kyle Genova, Pratul Srinivasan, Howard Zhou,
  Jonathan Barron, Ricardo Martin-Brualla, Noah Snavely, and Thomas Funkhouser.
\newblock {IBRNet: Learning Multi-View Image-Based Rendering}.
\newblock In {\em CVPR}, 2021.

\bibitem{Wang21arxiv_NeRFminusminus}
Zirui Wang, Shangzhe Wu, Weidi Xie, Min Chen, and Victor~Adrian Prisacariu.
\newblock {NeRF--: Neural Radiance Fields Without Known Camera Parameters}.
\newblock {\em https://arxiv.org/abs/2102.07064}, 2021.

\bibitem{Wu16nips_3d_gan}
Jiajun Wu, Chengkai Zhang, Tianfan Xue, William~T Freeman, and Joshua~B
  Tenenbaum.
\newblock Learning a probabilistic latent space of object shapes via 3d
  generative adversarial modeling.
\newblock In {\em NeurIPS}, 2016.

\bibitem{Wu20cvpr_symmetry}
Shangzhe Wu, Christian Rupprecht, and Andrea Vedaldi.
\newblock Unsupervised learning of probably symmetric deformable 3d objects
  from images in the wild.
\newblock In {\em CVPR}, 2020.

\bibitem{Wu15cvpr_3d_shapenet}
Zhirong Wu, Shuran Song, Aditya Khosla, Fisher Yu, Linguang Zhang, Xiaoou Tang,
  , and Jianxiong Xiao.
\newblock 3d shapenets: A deep representation for volumetric shapes.
\newblock In {\em CVPR}, 2015.

\bibitem{Xiang14wacv_pascal3d}
Yu Xiang, Roozbeh Mottaghi, and Silvio Savarese.
\newblock Beyond pascal: A benchmark for 3d object detection in the wild.
\newblock In {\em IEEE Winter Conference on Applications of Computer Vision
  (WACV)}, 2014.

\bibitem{ye2021shelf}
Yufei Ye, Shubham Tulsiani, and Abhinav Gupta.
\newblock Shelf-supervised mesh prediction in the wild.
\newblock In {\em Proceedings of the IEEE/CVF Conference on Computer Vision and
  Pattern Recognition}, pages 8843--8852, 2021.

\bibitem{YenChen20iros_iNeRF}
Lin Yen-Chen, Pete Florence, Jonathan Barron, Alberto Rodriguez, Phillip Isola,
  and Tsung-Yi Lin.
\newblock {iNeRF: Inverting Neural Radiance Fields for Pose Estimation}.
\newblock In {\em IROS}, 2021.

\bibitem{Yu21cvpr_pixelNeRF}
Alex Yu, Vickie Ye, Matthew Tancik, and Angjoo Kanazawa.
\newblock {pixelNeRF: Neural Radiance Fields from One or Few Images}.
\newblock In {\em CVPR}, 2021.

\bibitem{Zhi21iccv_SemanticNeRF}
Shuaifeng Zhi, Tristan Laidlow, Stefan Leutenegger, and Andrew Davison.
\newblock {In-Place Scene Labelling and Understanding with Implicit Scene
  Representation}.
\newblock In {\em ICCV}, October 2021.

\bibitem{Zuffi19iccv_safari}
Silvia Zuffi, Angjoo Kanazawa, Tanya Berger-Wolf, and Michael~J. Black.
\newblock Three-d safari: Learning to estimate zebra pose, shape, and texture
  from images "in the wild".
\newblock In {\em ICCV}, 2019.

\end{thebibliography}
}
\newpage
\appendix

\section{Evaluation Details}

In this section we clarify the implementation details presented in the paper. 

\begin{figure*}[t]
\centering
\includegraphics[width=0.85\linewidth]{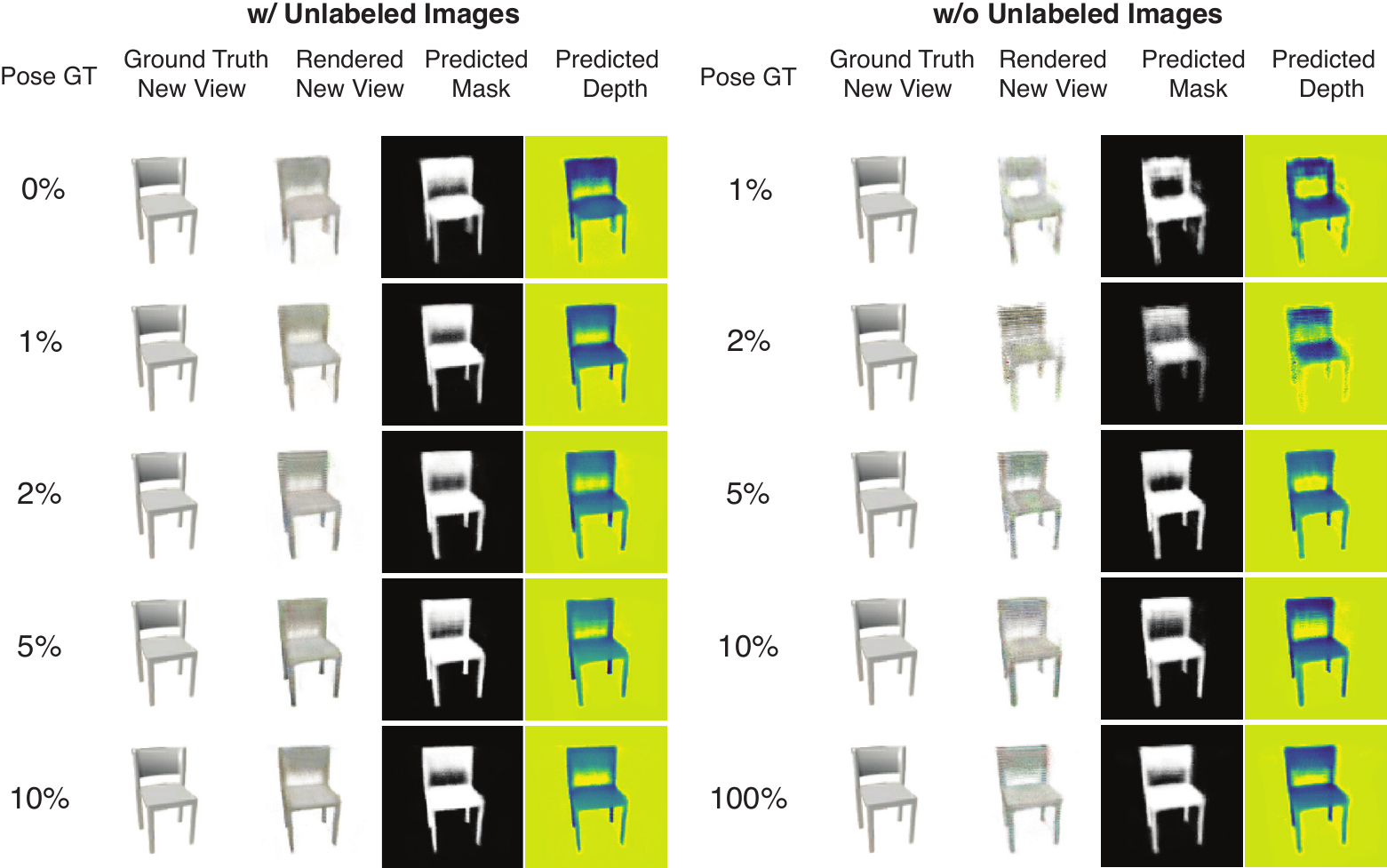}\
\caption{Visualization of our results with different levels of supervision on ShapeNet 3D-R2N2 \cite{choy20163d} chair. We find out that using as little as $1\%$ posed data, we can leverage the remaining $99\%$ unlabeled data to achieve results that are comparable to fully supervision (100\%) of pose.}
\label{fig:shapenet weak supervised chair images}
\end{figure*}

\textbf{Weakly-Supervision}. The results demonstrated in Table 1, Figure 3 in the main paper and  Figure~\ref{fig:shapenet weak supervised chair images}, are evaluated on ShapeNet 3D-R2N2 dataset~\cite{choy20163d}. In those experiments, single-view images are used as input. In the weakly supervised setup, a small fraction of those single-view images are provided with ground-truth camera pose. And we first pre-train the model with images that have known camera pose. Then we fine-tune using both posed images as well as images with unknown camera pose.

As shown in Figure~\ref{fig:shapenet weak supervised chair images}, we inspect our model under different levels of pose ground truth supervision (with single-view input). We show that our model by using only $1\%$ of posed data and leveraging the other $99\%$ of unlabeled data, could achieve results comparable to a fully supervised setting, as well as outperform training without unlabeled data.

\textbf{Unsupervised setting}. In the unsupervised setup, non of the images is provided with ground-truth camera pose. In these experiments, camera pose is predicted from the input single-view image. And during training, our model does not have access to ground-truth camera pose.  The results of our model with no pose supervision on ShapeNet 3D-R2N2 dataset~\cite{choy20163d} are demonstrated in Table 2, Table 3, Fig 4, Fig 5. The unsupervised results on Open Images dataset~\cite{kuznetsova2020open,Benenson19cvpr} are shown in Fig 6, Fig 7. 

\textbf{Multi-view Supervision}. We also evaluate our model in a setting where the pose is given as input, and the model has access to multiple pose views of an object during training (during evaluation it is still provided with a single-view images).

In the Table~\ref{table: multi/singl-view supervision} below, We summarize our results for these different settings. 

\begin{table*}[t]
\centering
\begin{tabular}{l|c|cc|cc|cc}
\hline
 & Category & \multicolumn{2}{c|}{Chair} & \multicolumn{2}{c|}{Car} & \multicolumn{2}{c}{Airplane} \\ \hline
View & Pose & PSNR & SSIM & PSNR & SSIM & PSNR & SSIM \\ \hline
multi- & input & 20.8 &0.83 & 22.2  &0.86  &25.8  &0.94  \\ \cline{1-2}
\multirow{3}{*}{single-} & input & 19.3 & 0.80 & 21.0 & 0.85 & 25.4 & 0.93 \\ \cline{2-2}
 & fully-supervised & 19.0 & 0.77 & 20.1 & 0.83 & 25.2 & 0.93 \\ \cline{2-2}
 & weakly-supervised & 18.6 & 0.77 & 19.5 & 0.82 & 24.8 & 0.92 \\ 
  & unsupervised &16.4  &0.72 &17.2  &0.74  &22.2 &0.87  \\ 
 
 \hline
 
 \end{tabular}
\label{table: multi/singl-view supervision}
\caption{Comparison between model with multi-view v.s. single-view supervision, and whether pose is directly given as input or predict from the single-view image input given different levels of supervision (fully-supervised 100\% v.s. weakly-supervised 1\% v.s. unsupervised 0\%) on ShapeNet 3D-R2N2 dataset~\cite{choy20163d} chair, car and airplane.}

\end{table*}

\section{Comparison to Baselines}
As described in the paper, we compare our method with multiple baselines: PrGAN~\cite{gadelha20173d}, HoloGAN~\cite{nguyen2019hologan} and shelf-supervised mesh prediction by Ye et al~\cite{ye2021shelf}.  We use the open-source implementation of PrGAN and HoloGAN by Ye et al~\cite{ye2021shelf}. HoloGAN is added with a function that maps the volumetric feature to occupancy using re-projection consistency with foreground masks. Each model only takes a single-view input and renders new views. These models are also trained with single-view unposed images using meta-supervision. Our model outperforms these approaches to a large extent, as demonstrated in Table 3 and Figure 6 in the main paper. 

Meanwhile, we also implement pixelnerf~\cite{Yu21cvpr_pixelNeRF}, which uses a spatial encoder to generate image feature as local embedding to the decoder, while model uses global latent code instead. For the local embedding, pose are required as inputs to project rays onto the image features during rendering, we find training such model with a spatial encoder without pose given is very difficult and the results of using local embedding are much worse than those from our models using global latent code.

\end{document}